\documentclass[journal,a4paper]{IEEEtran} %

\usepackage[
  top=2.5cm,       %
  bottom=2.5cm,       %
  left=1.47cm,         %
  right=1.47cm,        %
]{geometry}

\usepackage[top=2.5cm, bottom=2.5cm, left=1.47cm, right=1.47cm]{geometry}

\usepackage[utf8]{inputenc} 
\usepackage[T1]{fontenc}
\usepackage{url}
\usepackage{ifthen}
\usepackage{cite}
\usepackage[cmex10]{amsmath} %

\usepackage{amsthm}
\usepackage{amsfonts}
\usepackage{amssymb}
\usepackage[capitalize]{cleveref}
\usepackage{enumitem}
\usepackage{xspace}
\usepackage{xcolor}
\usepackage{balance}

\usepackage{algorithmic}
\usepackage{algorithm}
\usepackage{array}
\usepackage[caption=false,font=normalsize,labelfont=sf,textfont=sf]{subfig}
\usepackage{textcomp}
\usepackage{stfloats}
\usepackage{url}
\usepackage{verbatim}
\usepackage{graphicx}
\usepackage{cite}
\usepackage[normalem]{ulem}
\usepackage{pgfplots}
\pgfplotsset{compat=newest}
\usepackage{tikz}
\usetikzlibrary{backgrounds,calc,shadings,shapes.arrows,shapes.symbols,shadows,fit,positioning,topaths,hobby}

\newcommand{\client}{\ensuremath{i}}
\newcommand{\nclients}{\ensuremath{n}}
\newcommand{\ncol}{\ensuremath{z}}
\newcommand{\nbyz}{\ensuremath{b}}
\newcommand{\col}{\ensuremath{\mathcal{T}}}
\newcommand{\clienttmp}{\ensuremath{j}}
\newcommand{\clienttmptwo}{\ensuremath{\ell}}
\newcommand{\gi}[1][\client]{\ensuremath{\mathbf{g}_{#1}}}
\newcommand{\ssgi}[1][x]{\ensuremath{f_\client(#1)}}
\newcommand{\ssgj}[1][x]{\ensuremath{f_\clienttmp(#1)}}
\newcommand{\ssgk}[1][x]{\ensuremath{f_\clienttmptwo(#1)}}
\newcommand{\ssdistjk}[1][x]{\ensuremath{h_{\clienttmp, \clienttmptwo}(#1)}}
\newcommand{\roundpad}{\ensuremath{\mathbf{m}_\clienttmp}}
\newcommand{\sharegij}{\ensuremath{[\mathbf{g}_\client]_\clienttmp}}
\newcommand{\sharegji}{\ensuremath{[\mathbf{g}_\clienttmp]_\client}}
\newcommand{\sharegki}{\ensuremath{[\mathbf{g}_\clienttmptwo]_\client}}
\newcommand{\paddedsharegki}{\ensuremath{[\mathbf{g}_\clienttmptwo + \roundpad]_\client}}

\newcommand{\sharedistjk}[1][\client]{\ensuremath{[d_{\clienttmp, \clienttmptwo}]_{#1}}}
\newcommand{\sharedistnnmjk}[1][\client]{\ensuremath{[d^\Sigma_{\clienttmp, \clienttmptwo}]_{#1}}}

\newcommand{\ridx}{\ensuremath{t}}
\newcommand{\rit}[1][\ridx]{\ensuremath{\mathbf{r}_{\client, #1}}}

\newcommand{\ssk}[1][\ridx]{\ensuremath{k}}

\newcommand{\ssnnj}[1][x]{\ensuremath{g_\clienttmp(#1)}}
\newcommand{\sit}[1][\ridx]{\ensuremath{\mathbf{s}_{\client, #1}}}

\newcommand{\sharennji}{[\mathbf{g}^{\Sigma}_\clienttmp]_\client}

\newcommand{\define}{\ensuremath{\triangleq}}
\newcommand{\eval}[1][\clienttmp]{\ensuremath{\alpha_{#1}}}
\newcommand{\distij}{\ensuremath{d_{\client, \clienttmp}}}
\newcommand{\distjk}[1][\clienttmptwo]{\ensuremath{d_{\clienttmp, #1}}}
\newcommand{\distkj}[1][\clienttmptwo]{\ensuremath{d_{#1, \clienttmp}}}
\newcommand{\distnnmjk}[1][\clienttmptwo]{\ensuremath{d^\Sigma_{\clienttmp, #1}}}
\newcommand{\Fq}{\ensuremath{\mathbb{F}_q}}
\newcommand{\dimension}{\ensuremath{d}}
\newcommand{\norm}[1]{\Vert #1 \Vert}

\newcommand{\tmpvec}{\mathbf{g}_\client}
\newcommand{\tmpvecsorted}[1][\clienttmp]{\mathbf{g}_{\client, #1}}
\newcommand{\tmpnnm}[1][\client]{\mathbf{g}^{\Sigma}_{#1}}
\newcommand{\nnms}[1][\client]{\mathcal{N}_{#1}}
\newcommand{\gnnm}[1][\client]{\mathbf{g}^{\Sigma}_{#1}}
\newcommand{\obsi}[1][\client]{\ensuremath{\mathcal{O}_{#1}}}
\newcommand{\obsf}{\ensuremath{\mathcal{O}_F}}
\newcommand{\obsset}[1][\col]{\ensuremath{\mathcal{O}_{#1}}}

\newcommand{\condmutinf}[3]{\mathrm{I}\left(#1; #2 \vert #3\right)}

\newcommand{\data}[1][\client]{\mathcal{D}_{#1}}
\newcommand{\tmpclientset}{\ensuremath{\mathcal{C}}}

\newcommand{\distnnmij}{\ensuremath{d^\Sigma_{\client, \clienttmp}}}
\newcommand{\aggres}{\ensuremath{\mathbf{g}^\Sigma}}
\newcommand{\aggrest}{\ensuremath{\mathbf{g}^\Sigma_{(t)}}}
\newcommand{\robustaggop}{\ensuremath{\mathrm{R}}}
\newcommand{\quantizer}{\ensuremath{\mathrm{Q}}}
\newcommand{\nnmop}{\ensuremath{\mathrm{NNM}}}

\newcommand{\robustness}{\ensuremath{\kappa}}
\newcommand{\honest}{\ensuremath{\mathcal{H}}}

\newcommand{\krum}{Krum\xspace}

\newcommand{\partition}{\ensuremath{m}}

\newcommand{\iter}{\ensuremath{t}}
\newcommand{\model}[1][\iter]{\mathbf{w}_{#1}}
\newcommand{\loss}{F}
\newcommand{\git}{\mathbf{g}_\client^{(\iter)}}
\newcommand{\githat}{\hat{\mathbf{g}}_\client^{(\iter)}}
\newcommand{\seed}{\ensuremath{r}}
\newcommand{\githatentry}[1][\seed]{\hat{\mathbf{g}}_{\client, #1}^{(\iter)}}
\newcommand{\dimidx}{\ensuremath{u}}
\newcommand{\gitentry}{{g}_{\client, \dimidx}^{(\iter)}}

\newcommand{\clientstar}{\ensuremath{i^\star}}
\newcommand{\tmpclientstarset}{\ensuremath{\mathcal{C}^\star}}
\newcommand{\score}[1][\client]{\ensuremath{\vartheta_\client}}

\newcommand{\lr}{\ensuremath{\eta}}

\newcommand{\nseeds}{\ensuremath{R}}
\newcommand{\dev}{\ensuremath{\mu}}
\newcommand{\perturb}{\ensuremath{\mathbf{z}_r^\iter}}

\newcommand{\infmatrix}[1][\client]{\mathbf{U}}

\newcommand{\cwmatrix}[1][\client]{\mathbf{C}}

\newcommand{\rerandpoly}[1][x]{p(#1)}
\newcommand{\rerandrand}[1][x]{\nu_{#1}}
\newcommand{\quants}{\ensuremath{\mu}}

\begin{document}
\title{Private Aggregation for Byzantine-Resilient Heterogeneous Federated Learning}

\author{%
 \IEEEauthorblockN{Maximilian Egger and Rawad Bitar}\\ %
\IEEEauthorblockA{School of Information, Computation and Technology, Technical University of Munich, Germany\\
                   Email: \{maximilian.egger, rawad.bitar\}@tum.de}%
\thanks{This project is funded by DFG (German Research Foundation) projects under Grant Agreement Nos. BI 2492/1-1 and WA 3907/7-1.}
\vspace{-0.7cm}
}

\maketitle

\begin{abstract}
   Ensuring resilience to Byzantine clients while maintaining the privacy of the clients' data is a fundamental challenge in federated learning (FL). When the clients' data is homogeneous, suitable countermeasures were studied from an information-theoretic perspective utilizing secure aggregation techniques while ensuring robust aggregation of the clients' gradients. However, the countermeasures used fail when the clients' data is heterogeneous. Suitable pre-processing techniques, such as nearest neighbor mixing, were recently shown to enhance the performance of those countermeasures in the heterogeneous setting. Nevertheless, those pre-processing techniques cannot be applied with the introduced privacy-preserving mechanisms. 
   
   We propose a multi-stage method encompassing a careful co-design of verifiable secret sharing, secure aggregation, and a tailored symmetric private information retrieval scheme to achieve information-theoretic privacy guarantees and Byzantine resilience under data heterogeneity. We evaluate the effectiveness of our scheme on a variety of attacks and show how it outperforms the previously known techniques. Since the communication overhead of secure aggregation is non-negligible, we investigate the interplay with zero-order estimation methods that reduce the communication cost in state-of-the-art FL tasks and thereby make private aggregation scalable.
\end{abstract}

\theoremstyle{plain}
\newtheorem{corollary}{Corollary}%
\newtheorem{proposition}{Proposition}%
\newtheorem{theorem}{Theorem}%
\newtheorem{lemma}{Lemma}%
\newtheorem{claim}{Claim}%

\newtheorem{definition}{Definition}%
\newtheorem{construction}{Construction}%
\newtheorem{example}{Example}%
\newtheorem{approach}{Approach}%
\newtheorem{property}{Property}%
\newtheorem{remark}{Remark}
\newtheorem{assumption}{Assumption}

\section{Introduction}

Federated learning (FL) is a widely studied machine learning paradigm \cite{mcmahan2017communication} with contradicting challenges. Clients use their private data to contribute local gradients to a collaborative and iterative learning process centrally orchestrated by a federator. The underlying optimization principles often rely on (stochastic) gradient descent. Maintaining the privacy of the clients' data and ensuring resilience against Byzantine clients deliberately trying to corrupt the learning process are two major challenges in FL. Many countermeasures have been studied extensively in the literature, cf. the survey \cite{yin2021comprehensive} for privacy, and \cite{rodriguez2023survey} for security, i.e., resilience against Byzantine behavior. Secure aggregation, introduced in~\cite{bonawitz2017practical}, is one approach to ensure the individual clients' privacy and was further studied in, e.g., \cite{bell2020secure,so2021turbo,kadhe2020fastsecagg,so2022lightsecagg,jahani2022swiftagg+,schlegel2023coded,sami2023secure,egger2024private} to reduce the cost of communication in various system models. Byzantine-resilient FL was studied in, e.g., \cite{blanchard2017machine,allen2021byzantine,chen2017distributed,el2021collaborative,karimireddy2021learning,karimireddy2022byzantinerobust,ozfatura2023byzantines}. Despite being of similar importance, the objectives of privacy and security are contradictory in nature. While privacy requires hiding sensitive information about individual clients from other clients and the federator, security requires the federator to learn and compare statistics from the individual clients' contributions to distinguish between honest and Byzantine behavior to sort out corrupt contributions. The authors of \cite{so2020byzantine} first investigated how to simultaneously achieve privacy and security from an information-theoretic perspective using tools from secret sharing \cite{shamir1979how}. The paper uses Multi-\krum \cite{blanchard2017machine}, which is a well-known robust aggregation rule. However, as a result of using Multi-\krum, the privacy measure is relaxed to leaking distances between the local gradients of the clients to the federator. To reduce the communication cost, an improved version based on ramp secret sharing~\cite{mceliece1981sharing} was proposed in \cite{jahani2023byzsecagg}. A general framework for private and robust aggregation was proposed in \cite{hou2024priroagg} with two instantiations using the aggregation rules Robust Federated Averaging \cite{pillutla2022robust} and Robust Learning Rate \cite{ozdayi2021defending}. The former is less robust than Multi-\krum \cite{pillutla2022robust}, and the latter is designed mainly against backdoor attacks. All those methods \cite{so2020byzantine,jahani2023byzsecagg,hou2024priroagg} rely on cryptographic primitives, hence reducing the privacy guarantees to computational privacy relying on a hard mathematical problem. The authors of \cite{xia2024byzantine} were the first to introduce an end-to-end information-theoretic protocol for robust and private FL, combining the robust aggregation of FLTrust \cite{cao2020fltrust} with information-theoretic verifiable secret sharing (ITVSS) \cite{ben2019completeness}. The cost of ensuring the strongest possible privacy is an increased communication cost. The communication cost was further reduced by introducing a trusted third party~\cite{xia2024lobyitfl}. Different from the other proposed methods, the use of FLTrust requires the federator to have access to a public dataset. %

Most robust aggregation schemes (like \krum and Multi-\krum \cite{blanchard2017machine}) rely on the data of the clients being homogeneous, i.e., the clients' data is drawn from the same underlying mother distribution. However, this assumption is not satisfied in realistic FL scenarios where the clients generate their own data, potentially from different distributions. It was shown that robust aggregation is significantly more challenging under data heterogeneity \cite{el2021collaborative,karimireddy2022byzantinerobust}, further underlined by \cite{charikar2017learning,liu2021approximate}.

Motivated by robust aggregation for heterogeneous data, the authors of \cite{allouah2023fixing} introduced a technique termed Nearest Neighbor Mixing (NNM) to pre-process the local gradients of the clients %
before applying off-the-shelf robust aggregation. They show that NNM renders most robust aggregation rules optimal under data heterogeneity. On a high level, the clients' gradients are first mixed, e.g., averaged, among their nearest neighbors and subsequently fed into existing robust aggregation rules like \krum. Although promising, incorporating the composition of such pre-processing and robust aggregation into a privacy-preserving framework under information-theoretic guarantees is challenging due to the computation of nearest neighbors, the gradient mixing operations, and the application of robust aggregation. Information-theoretic privacy requires that neither any $\ncol$ colluding clients nor the federator should learn anything about other clients' gradients or their nearest neighbors through any observations, while ensuring robustness of the protocol against $\nbyz$ Byzantine clients.

\textbf{Contributions.} We introduce a novel protocol that brings optimality and privacy to robust aggregation in heterogeneous FL. We only allow the leakage of the distances between the clients' gradients and their nearest neighbor mixtures to the federator, who remains unaware of any individual client's gradient or any aggregation of a subset of clients' gradients apart from the final computation result, strictly necessary to advance the training process. We formally define and prove the desired information-theoretic privacy guarantees, and prove the robustness of the protocol when the number of clients $\nclients$ satisfies $\nclients > \max\{3\nbyz, 2(\ncol+\nbyz)\}$. We derive and compare the communication cost of our scheme to BREA \cite{so2020byzantine} and ByzSecAgg \cite{jahani2023byzsecagg} (which both leak pair-wise gradient distances but rely on cryptographic measures), and show through numerical experiments on a variety of state-of-the-art attacks significant improvements in heterogeneous regimes, both for \krum and Multi-\krum with stochastic quantization necessary to provide information theoretic privacy guarantees. BREA and ByzSecAgg use Multi-\krum to prevent individual gradient leakage, which lacks robustness guarantees. By incorporating NNM, our method safely uses both Multi-\krum and \krum, leveraging the latter’s known theoretical guarantees \cite{allouah2023fixing}. To reduce the communication cost of private aggregation, we use zero-order gradient estimation, which is especially helpful in tasks such as fine-tuning Large Language Models (LLMS) \cite{malladi2023fine}. We build on the non-private robust aggregation schemes recently introduced in \cite{neto2024communication,egger2025byzantine}. %
We test our approach on fine-tuning LLMs using zero-order optimization and show that, while ensuring privacy, we achieve state-of-the-art performances and significantly reduce the communication cost of private aggregation.

\section{Preliminaries and System Model}

\textbf{Federated learning.} We consider a system of $\nclients$ clients  $\client \in [\nclients] \define \{1, \cdots, \nclients\}$, each possessing a private dataset $\data$. Collaboratively, they iteratively train a machine learning model centrally orchestrated by a federator. Starting from a common model $\model[0]$ at iteration $\iter = 0$, at each iteration $\iter$, the clients compute a local gradient\footnote{Local iterations are possible and fit within our proposed method without any further adaptations. The quantity of interest, in this case, is the aggregation of gradients from multiple local iterations.} $\git \define \nabla_{\model[]} \loss(\model, \data) \in \mathbb{R}^{\dimension}$ based on a given loss function $\loss$, the current model $\model$, and the private data $\data$. We will also consider a zero-order gradient estimate $\githat$ that drastically reduces the communication cost.
In the sequel, we will use the notation $\git$ for the gradient; however, all our concepts apply for gradient estimates $\githat$\!.

The gradients are then quantized into a finite field $\Fq$ according to a stochastic quantizer $\quantizer$ that operates on the gradient entries $\gitentry, \dimidx \in [\dimension],$ and preserves the expectation of the gradient, i.e., $\mathbb{E}[\quantizer(\git)] = \git$. We use the stochastic quantization from \cite{gupta2015deep,croci2022stochastic} with $\quants$ quantization levels and omit the details for brevity.
The field size $q$ is chosen large enough to avoid overflows in the gradient aggregation, i.e., $q \geq 2\dimension (\nclients-\nbyz)^2\quants^2$. %
The quantized gradients\footnote{The protocol can operate equivalently on the clients' local models, i.e., models and gradients can be used interchangeably.} $\quantizer(\gi)$ are sent to the federator and aggregated according to an aggregation rule $\robustaggop$ to obtain a global gradient $\aggrest \define \robustaggop(\{\quantizer(\git)\}_{\client \in [\nclients]})$. The simplest example of $\robustaggop$ is averaging the gradients, i.e., $\robustaggop_{\text{avg}} = \frac{\sum_{i=1}^{n}\quantizer(\git)}{n}$. The model is updated as $\model[\iter+1] = \model[\iter] - \lr \robustaggop(\{\quantizer(\git)\}_{\client \in [\nclients]})$, where $\lr$ is the learning rate. The iterative model update is repeated until meeting certain convergence criteria. Since the protocol is independently executed at every iteration, we focus on one iteration and omit the index $\iter$. In the remainder of the paper, we refer to the quantized gradients $\quantizer(\gi)$ as $\gi$ for ease of notation.

\textbf{Adversarial model.} We assume the existence of at most $\nbyz$ Byzantine clients that act in an arbitrary manner to corrupt the learning process. We assume here the strongest possible notion, where Byzantine clients can collaborate and have full knowledge about the protocol and all clients' data. This is a very strong model, considering that our privacy guarantees will avoid any information leakage among clients. In addition, we assume that at most $\ncol$ clients collude to compromise the other clients' privacy, e.g., to infer their private data. Further, the federator should only learn the aggregation result $\aggres$, which is the only necessary information for the learning process. 

\textbf{Byzantine resilience.} It is shown~\cite{blanchard2017machine} that, without any defense mechanism, one Byzantine client can corrupt the whole learning algorithm. To mitigate the effect of Byzantine clients, several robust aggregation rules have been proposed, e.g., coordinate-wise trimmed median \cite{yin2018byzantine}, \krum and Multi-\krum \cite{blanchard2017machine}. To capture the properties of a good robust aggregation rule, the following definition was introduced, cf. \cite{allouah2023fixing}. %

\begin{definition}[$(\nbyz, \robustness)$-Robust Aggregation] Let $\robustness \geq 0$ and $\nbyz < \nclients/2$. For inputs $\mathbf{g}_1, \cdots, \mathbf{g}_\nclients$ and any set $\honest \subset [\nclients]$ of size $\vert \honest \vert = (\nclients-\nbyz)$, an aggregation rule $\robustaggop$ is $(\nbyz, \robustness)$-robust if
\begin{equation*}
    \norm{\robustaggop(\mathbf{g}_1, \cdots, \mathbf{g}_\nclients) - \bar{\mathbf{g}}_\honest}^2 \leq  \sum_{\client \in \honest} \frac{\kappa}{\vert \honest \vert} \norm{\mathbf{g}_\client - \bar{\mathbf{g}}_\honest}^2, \vspace{-.3cm}
\end{equation*}
where $\bar{\mathbf{g}}_\honest = \frac{1}{\vert \honest \vert} \sum_{\client \in \honest} \mathbf{g}_\client$.
\end{definition}

In \cite{allouah2023fixing}, the authors show that existing robust aggregation rules, despite providing good robustness, do not lead to desirable convergence guarantees of learning algorithms under data heterogeneity. 
To remedy this drawback, \cite{allouah2023fixing} proposes NNM pre-processing that, when composed with an $(\nbyz, \robustness)$ robust aggregation rule $\robustaggop$, improves its robustness and is empirically shown to improve convergence of several learning algorithms. The convergence is even proven to be optimal for non-stochastic settings, cf. \cite[Theorem 1]{allouah2023fixing}. 
\cref{def:nnm} is a slightly modified version of NNM, without normalization, that is compatible with finite field operations. %
\begin{definition}[Nearest Neighbor Mixing] \label{def:nnm}
    Let $(\mathbf{g}_1, \cdots, \mathbf{g}_\nclients)$ be a set of vectors from $\nclients$ clients, and $\nbyz$ the maximum number of Byzantine clients. %
    For each $\mathbf{g}_\client$, let $(\tmpvecsorted[1], \cdots, \tmpvecsorted[\nclients])$ be the ordered vectors such that $\norm{\tmpvecsorted[1] - \tmpvec}_2 \leq \cdots \leq \norm{\tmpvecsorted[\nclients] - \tmpvec}_2$. NNM outputs $(\tmpnnm[1], \cdots, \tmpnnm[\nclients])$, where\footnote{The original NNM computes the average of the nearest neighbors. However, since the robust aggregation $\robustaggop$ is distance-based in our case, there is no difference in operating on the sum of gradients instead of their average.}
        $\tmpnnm = \sum_{\clienttmp = 1}^{\nclients-\nbyz} \tmpvecsorted.$
\end{definition}

Note that, although used in \cite{so2020byzantine,jahani2023byzsecagg}, Multi-\krum is only empirically shown to be Byzantine resilient; however, there is no theoretical guarantee on the value of $\robustness$ that it attains.

\textbf{Privacy Guarantees. }
To compute nearest neighbor mixtures for the gradient of each client $\client$ followed by a robust aggregation method, both the clients and the federator will exchange multiple messages with each other. We denote by $\obsi$ all observations made by client $\client$, and by $\obsf$ all observations made by the federator, and use the following set notation $\obsset[\col] \define \{\obsi\}_{\client \in \col}$ for a client set $\col\subset [\nclients]$. Let further $\aggres \define \robustaggop \circ \nnmop(\{\gi\}_{\client \in [\nclients]})$ be the result of the composition of NNM and the robust aggregation of gradients. For random variables $X,Y$ and $Z$, let $\condmutinf{X}{Y}{Z}$ denote the mutual information between $X$ and $Y$ conditioned on $Z$. With a slight abuse of notation, we impose the following privacy objectives:
\begin{definition}[Privacy from Colluding Clients] \label{def:privacy_clients} For each client $\client \in [\nclients]$, its gradient $\gi$ and the identity of its nearest neighbors $\nnms[\client]$ should remain private from any set $\col \subset [\nclients] \setminus \{\client\}, \vert \col \vert \leq \ncol$, of colluding clients, i.e.,
\begin{align*}
    \condmutinf{\gi, \nnms[\client]}{\obsset[\col]}{\{\gi[\clienttmp]\}_{\clienttmp \in \col}, \aggres} = 0, \forall \client \in [\nclients].
\end{align*}
\end{definition}
Our scheme achieves a slightly weaker privacy guarantee (cf. \cref{thm:properties}), but can be modified to attain \cref{def:privacy_clients}.

Let $\distij \define \norm{\gi - \gi[\clienttmp]}_2^2$ denote the distance between the gradients of client $\client$ and $\clienttmp$, and $\distnnmij\define \norm{\gnnm[\client] - \gnnm[\clienttmp]}_2^2$ denote the distance between their nearest neighbor mixtures, then we define the following private measure.
\begin{definition}[Privacy from the Federator] \label{def:privacy_federator} The federator learns no information about the clients' gradients beyond the final aggregation $\aggres$ and the pair-wise distances between the clients' gradients and their nearest neighbor mixtures, i.e.,
\begin{align*}
        \condmutinf{\{\gi\}_{\client \in [\nclients]}}{\obsf}{\{\distij, \distnnmij\}_{\client, \clienttmp \in [\nclients]}, \aggres} = 0.
\end{align*}
\end{definition}
Hence, in addition to $\aggres$, we allow the distances between the clients' gradients and their nearest neighbor mixtures to be leaked to the federator. While those distances are potentially revealed, the corresponding gradients remain private. The clients, on the other hand, are not allowed to learn the distances, the gradients, nor the mixtures of gradients of other clients beyond their gradients and potential correlations that are impossible to hide, and the final aggregation necessary to advance the training. We note that, although we focus on information-theoretic privacy measures, combinations with other privacy measures are possible and might be useful in many cases \cite{ngo2024secure}. Collusion between clients and the federator is not allowed in this model.

\section{Main Result}

We seek a protocol that executes the composition of NNM with a robust aggregation rule while respecting the privacy guarantees. This is challenging since: i) all individual gradients and their mixtures should remain private from the federator and other clients, and ii) a client's nearest neighbors should remain private from the other clients. To overcome these challenges, we propose a protocol consisting of secret sharing operations, a private function retrieval with Byzantine resilience from MDS-coded data, and re-encoding of padded mixtures followed by standard procedures. To avoid introducing a trusted third party, we assume shared randomness between the clients, unknown to the federator. This can be instantiated through a one-time use of a trusted random number generator.

We state the main properties of our protocol: privacy, Byzantine resilience, and communication cost. The protocol is described in the following section. For the theorem statement, let $\tmpclientstarset$ be the set of clients declared honest by the federator.

\begin{theorem} \label{thm:properties}
    Consider an FL algorithm with $\nclients$ clients out of which at most $\ncol$ collude to breach the privacy of other clients and at most $\nbyz$ are Byzantine, such that $\nclients > \max\{3\nbyz, 2(\ncol+\nbyz)\}$. The protocol introduced in \cref{sec:protocol} satisfies:    
    \begin{enumerate}
        \item The following formal privacy guarantees are ensured:\footnote{The conditioning on $\tmpclientstarset$ is analog to \cite{so2020byzantine,jahani2023byzsecagg} and can be relaxed by adding a round of private sum retrieval, mentioned in \cref{sec:protocol}, step 8).} $\condmutinf{\{\gi\}_{\client \in [\nclients]}}{\obsf}{\{\distij, \distnnmij\}_{\client, \clienttmp \in [\nclients]}, \aggres} = 0$ (cf. \cref{def:privacy_federator}) and $\condmutinf{\gi, \nnms[\client]}{\obsset[\col]}{\{\gi[\clienttmp]\}_{\clienttmp \in \col}, \aggres\!, \tmpclientstarset\!} = 0, \forall \client \in [\nclients]$.
    \item The protocol is corruption-resilient against at most $\nbyz$ clients. Using a distance-based $(\nbyz, \robustness)$-robust aggregation, our algorithm attains $(\nbyz, \frac{8\nbyz}{\nclients-\nbyz} (\robustness+1))$-robustness. %
    \item When using \krum or Multi-\krum as a robust aggregation rule, the per-user communication cost of our scheme is $\mathcal{O}(\dimension + \dimension \nclients^2 + \nclients^2)$, and $\mathcal{O}(\dimension \nclients^2 + \nclients^3)$ for the federator.
    \end{enumerate}
    
The proofs will be provided in an extended version. %

\end{theorem}

\section{Robust and Private Aggregation with NNM}\label{sec:protocol}

Our protocol is divided into multiple parts: 1) secret sharing gradients among clients, 2) private pair-wise distance computation by clients, 3) reconstruction of distances by the federator and selection of nearest neighbors, 4) preparation of gradient shares, 5) private sum retrieval, 6) re-encoding and sharing of private mixtures, 7) reconstruction of mixtures, and 8) robust aggregation. We detail the protocol in the following. Throughout the protocol, each client $\client \in [\nclients]$ is dedicated an evaluation point $\eval[\client]\in\mathbb{F}_q$. For clarity of exposition, we present the scheme with Shamir secret sharing as in \cite{shamir1979how} and note that the protocol can be extended to ramp secret sharing schemes \cite{mceliece1981sharing} to reduce the communication cost.

\begin{enumerate}[wide, labelwidth=!, labelindent=0pt]
\item \textbf{Sharing Among Clients:} 
Client $\client \in [\nclients]$ generates $\ncol$ random vectors $\rit[1],\dots,\rit[\ncol]$, uniformly at random from $\mathbb{F}_q^d$ and encodes its gradient $\gi$ into a secret sharing of the form
\begin{align*}
     \ssgi = \gi + x \rit[1] + \cdots + x^{\ncol} \rit[\ncol]
\end{align*}
and sends a share $\sharegij \define \ssgi[\protect{\eval[\clienttmp]}] \in \Fq^{\dimension}$ to each client $\clienttmp \in [\nclients] \setminus \client$, where $\eval[\clienttmp]$ is the evaluation point of client $\clienttmp$. To verify potentially corrupt secret sharing constructions by malicious clients\footnote{When the verification fails, the respective clients can either be excluded from the remainder of the protocol or the process can be restarted.}, we utilize the information-theoretic verifiable secret sharing scheme from \cite{ben2019completeness}, which requires that $\nclients > 3\nbyz$. The per-user communication cost for verifiable sharing is $\mathcal{O}(\dimension \nclients^2)$.

\item \textbf{Pair-wise Distance Computation:} Having evaluations from each client's gradient, client $\client$ computes a share of the pair-wise distance for each pair of clients $\clienttmp, \clienttmptwo \in [\nclients], \clienttmp < \clienttmptwo$. To avoid additional leakage resulting from multiplication on the shares, re-randomization is required. Therefore, we let clients share common randomness\footnote{Interactive approaches like in \cite{jahani2023byzsecagg} can remove the shared randomness assumption, but are outside the scope of this paper.}. The clients construct a scalar polynomial $\rerandpoly = x \rerandrand[1] + \cdots + x^{2\ncol} \rerandrand[2\ncol]$ (zero-coefficient equals $0$), where $\rerandrand[1], \cdots \rerandrand[2\ncol] \in \Fq$ are uniformly at random drawn from the shared source of randomness. Then, client $\client$ computes a share of $\distjk = \norm{\gi[\clienttmp] - \gi[\clienttmptwo]}_2^2$ as
\begin{align*}
    \sharedistjk \! = \! \norm{\sharegji \! - \! \sharegki}_2^2 \! + \! \rerandpoly[\protect{\eval[\client]}] \! = \! \norm{\ssgj[\protect{\eval[\client]}] \! - \! \ssgk[\protect{\eval[\client]}]}_2^2 \! + \! \rerandpoly[\protect{\eval[\client]}],
\end{align*}
which is an evaluation of a polynomial $\ssdistjk = \norm{\ssgj - \ssgk}_2^2 + \rerandpoly$ with degree $2\ncol$, where $\ssdistjk[0] = \distjk$. The field size $q$ is chosen large enough to avoid overflows in the computations. Analog to $\ssgi$, the polynomial $\ssdistjk$ corresponds to the encoding polynomial of a Reed-Solomon code. Hence, the federator can reconstruct the correct distances for all client pairs $\clienttmp, \clienttmptwo \in [\nclients]$ upon receiving $2(\ncol + \nbyz)+1$ evaluations through error-correction. The per-user communication complexity for transmitting the results of the distance computation is $\mathcal{O}(\nclients^2)$.

\item \textbf{Distance Reconstruction and Neighbor Selection:}
Receiving $2(\ncol + \nbyz)+1$ evaluations of $\ssdistjk$ for all $\clienttmp, \clienttmptwo$, the federator obtains the distances $\distjk$ and sorts them according to NNM in \cref{def:nnm}. Hence, for all $\clienttmp \in [\nclients]$, the federator has the indices of the $\nclients-\nbyz$ nearest neighbors $\nnms[\clienttmp]$ such that for any $\clienttmptwo \in \nnms[\clienttmp]$ and $\clienttmptwo^\prime \in [\nclients] \setminus \nnms[\clienttmp]$ it holds $\distjk[\clienttmptwo] \leq \distjk[\clienttmptwo^\prime]$. In contrast to the naive solution of broadcasting the set of all nearest neighbors $\{\nnms[\clienttmp]\}_{\clienttmp \in [\nclients]}$ to all clients for private aggregation of the nearest neighbors, we perform the following steps to hide the identity of $\nnms[\clienttmp]$ from all clients, including $\clienttmp$.%

\begin{figure}[!t]
    \centering
    \resizebox{.99\linewidth}{!}{\input{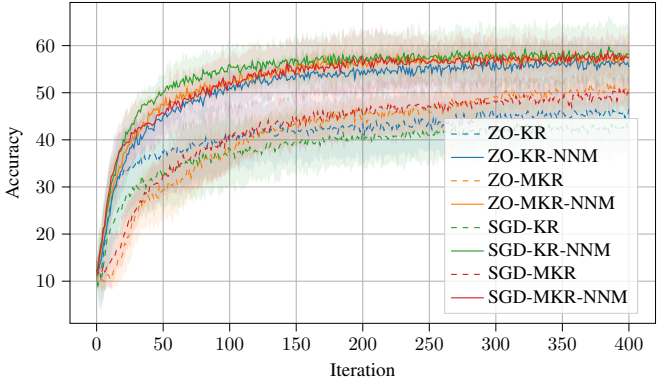}}
    \vspace{-.6cm}
    \caption{Averaged test accuracies over $5$ runs with different seeds for label flipping attack for different countermeasures and optimization methods with and without NNM. The performance of BREA \cite{so2020byzantine} and ByzSecAgg \cite{jahani2023byzsecagg} is shown as SGD-MKR. NNM significantly improves the Byzantine robustness. Shaded areas depict the standard deviations. See \cref{sec:experiments} for details. \vspace{-.4cm}}
    \label{fig:label_flipping}
\end{figure}

The two following steps are needed to privately compute the nearest neighbor mixture $\gnnm[\clienttmp]$ at the federator without revealing $\nnms[\clienttmp]$ to the clients or any partial aggregation of the gradients to the federator. We explain the next two steps for retrieving $\gnnm[\clienttmp]$ and note that those steps should be repeated for each $\clienttmp\in[\nclients]$.

\item \textbf{Preparing the Gradient Shares:} To hide the identity of the nearest neighbors $\nnms[\clienttmp]$, we will use tools from PIR. However, without further countermeasures on the coded data, any PIR procedure will result in the federator observing partial summations of the clients' gradients. To mitigate this effect, we utilize the clients' shared randomness. %
Each client $\client$ pads their secret shares $\sharegki, \clienttmptwo \in [\nclients]$ with a common random vector $\roundpad$ known to all clients but unknown to the federator, i.e., the client computes $\sharegki + \roundpad = \paddedsharegki$. Due to the polynomial structure, this corresponds to padding the polynomial coefficient that contains the private gradient $\gi[\clienttmptwo]$. In the following, the clients operate on the padded shares $\paddedsharegki$. Note that each client pads with the same random vector for a fixed NNM query of client $\clienttmp$.

\item \textbf{Private Sum Retrieval:} We use an extended symmetric PIR scheme to allow the federator to privately retrieve only the aggregation $\sum_{\clienttmptwo \in \nnms[\clienttmp]} \gi[\clienttmptwo] + \vert \nnms[\clienttmp] \vert \roundpad$ from the clients without revealing the identity of the clients in $\nnms[\clienttmp]$. For clarity of presentation, we omit the details of the PIR scheme. %
On a high level, the clients here can be seen as the servers in classical PIR, each storing codewords $\paddedsharegki$, $\client\in [\nclients]$ of Reed Solomon codes for each client $\clienttmptwo$'s gradient (analog to the files in coded PIR) in the form of secret shares. %
Carefully modifying the scheme in~\cite{tajeddine2019private} to return linear combinations of files, by querying at least $\ncol+2\nbyz+1$ clients, the federator can decode the padded sum $\gnnm[\clienttmp] + \vert \nnms[\clienttmp] \vert \roundpad$ of the nearest neighbors of client $\clienttmp$ even in the presence of $\nbyz$ Byzantine clients. The scaling $\vert \nnms[\clienttmp] \vert = \nclients-\nbyz$ is known and public. %
The communication per-client is in $\mathcal{O}(\dimension \nclients + \nclients^2)$, and for the federator in $\mathcal{O}(\dimension \nclients^2 + \nclients^3)$.

\item \textbf{Re-encoding and Sharing the Mixtures:} For all $\client, \clienttmp \in [\nclients]$, the federator sends to client $\client$ a share of the padded nearest neighbor mixture $\clienttmp \in [\nclients]$ according to the following encoding:
\begin{align*}
    \ssnnj = \gnnm[\clienttmp] + (\nclients-\nbyz) \roundpad + x \sit[1] + \cdots + x^{\ncol} \sit[\ncol],
\end{align*}
where $\sit, \ridx \in [\ncol],$ are drawn uniformly from $\Fq^\dimension$. The share $\ssnnj[\protect{\eval[\client]}]$ is transmitted to client $\client$. %
The communication cost of re-sharing is $\mathcal{O}(\dimension \nclients^2)$ for the federator, and $\mathcal{O}(\dimension \nclients)$ per client.

\item \textbf{Reconstructing the Mixtures:} Knowing the one-time pad $\roundpad$, for each nearest neighbor mixture of client $\clienttmp$, each client $\client \in [\nclients]$ can remove the padding and thus obtain a share $\sharennji$ of $\gnnm[\clienttmp]$ by computing $\sharennji \define \ssnnj[\protect{\eval[\client]}] - (\nclients-\nbyz) \roundpad$. Hence, each client $\client$ owns a share of the polynomial $\gnnm[\clienttmp] + x \sit[1] + \cdots + x^{\ncol} \sit[\ncol]$ for all $\clienttmp \in [\nclients]$. Note that this is similar to the starting point of existing schemes (such as \cite{so2020byzantine,jahani2023byzsecagg}) after the clients share their gradients amongst each other.

\begin{figure}[!t]
    \centering
    \resizebox{.99\linewidth}{!}{\input{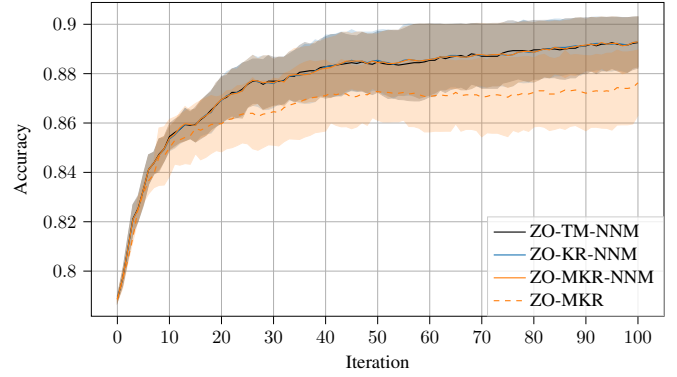}}
    \vspace{-.6cm}
    \caption{Averaged test accuracies over $3$ runs with different seeds for fine-tuning SST-2 with RoBERTa-large. The Byzantine clients conduct ALIE. ZO-TM-NNM is the straightforward extension of \cite{neto2024communication} to NNM without privacy. Both \krum and Multi-\krum can achieve state-of-the-art accuracies in this setting. See \cref{sec:experiments} for details.\vspace{-0.3cm}}
    \label{fig:fine_tuning}
\end{figure}

The re-encoding of the secret shares is strictly necessary to avoid any leakage of the nearest neighbor set to the clients that can happen by observing shares of the other clients' gradients and a corresponding share of the nearest neighbor mixture. %

\item \textbf{Private Robust Aggregation:} Any distance-based robust aggregation rule, e.g., \krum or Multi-\krum, can be computed on the shares $\sharennji$. Client $\client \in [\nclients]$ computes re-randomized shares of the pair-wise distances $\sharedistnnmjk$ between clients $\clienttmp, \clienttmptwo \in [\nclients]$, $\clienttmp < \clienttmptwo$ (equivalent to step 2, but on shares of $\gnnm[\client]$). The results are shared with the federator, decoding the distances\footnote{Note the distances are symmetric, and hence we set $\distjk \define \distkj$ for $\clienttmp > \clienttmptwo$.\vspace{.01cm}} $\distnnmjk, \clienttmp < \clienttmptwo,$ and selecting the presumably non-Byzantine clients to be aggregated, hereby referred to as $\tmpclientstarset$. For example, \krum selects the client $\clientstar$ with the lowest score, i.e., $\clientstar = \arg\min_{\clienttmp \in [\nclients]} \sum_{\clienttmptwo \in \tmpclientset_\clienttmp} \distnnmjk$, where $\tmpclientset_\clienttmp$ contains the closest $\nclients-\nbyz-2$ clients $\clienttmptwo$ to client $\clienttmp$ in terms of $\distnnmjk$. Hence, $\tmpclientstarset = \{\clientstar\}$. Multi-\krum iteratively selects the $\nclients-2\nbyz-3$ clients with the lowest scores as follows. Starting from an empty set $\tmpclientstarset = \emptyset$, for $\nclients-2\nbyz-3$ iterations, the chosen clients are updated as $\tmpclientstarset \gets \tmpclientstarset \cup \{\clientstar\}$, where $\clientstar = \arg\min_{\clienttmp \in [\nclients]\setminus \tmpclientstarset} \sum_{\clienttmptwo \in \tmpclientset_\clienttmp} \distnnmjk$, and $\tmpclientset_\clienttmp \subset [\nclients] \setminus \tmpclientstarset$ contains the $\nclients-\nbyz-\vert \tmpclientstarset \vert -2$ clients $\clienttmptwo$ from $[\nclients] \setminus \tmpclientstarset$ closest to client $\clienttmp$. %

The federator signals the clients which contributions (i.e., the set $\tmpclientstarset$) are to be aggregated. Honest clients aggregate and send the desired contributions using their owned shares, and Byzantine clients send a random vector of their choice. Using the error-correction capability of the Reed-Solomon code encoding the shares, in the presence of at most $\nbyz$ corrupt share aggregations, the federator can decode the desired sum\footnote{To hide the identities of the clients in $\tmpclientstarset$, this naive aggregation can be replaced by another round of private sum retrieval, similar to step 5).}. Depending on the aggregation rule, the federator decodes, e.g., a single $\gnnm[\clientstar]$ for a selected client $\clientstar$ (\krum), or a partial aggregation $\sum_{\clientstar \in \tmpclientstarset} \gnnm[\clientstar]$ (Multi-\krum), which is mapped back to $\mathbb{R}^\dimension$ according to the inverse quantization $\quantizer^{-1}(\cdot)$ and normalized by the number of clients $\vert \tmpclientstarset \vert$. The output of the aggregation reads $\robustaggop \circ \nnmop(\{\gnnm[\client]\}_{\client \in [\nclients]}) = \aggres$, where $\aggres = \frac{1}{\vert \tmpclientstarset \vert} \sum_{\clientstar \in \tmpclientstarset} \gnnm[\clientstar]$, and where $\vert \tmpclientstarset \vert = 1$ for \krum. The model is updated with an additional normalization of $\frac{1}{\nclients-\nbyz}$ for the NNM operation as $\model[\iter+1] = \model[\iter] - \frac{\lr}{(\nclients-\nbyz)} \quantizer^{-1}\left(\robustaggop \circ \nnmop(\{\gi\}_{\client \in [\nclients]}) \right) = \model[\iter] - \frac{\lr}{\vert \tmpclientstarset \vert (\nclients-\nbyz)} \quantizer^{-1}\left(\sum_{\clientstar \in \tmpclientstarset} \gnnm[\clientstar]\right)$. 
The per-user communication for sharing the pair-wise distances is $\mathcal{O}(\nclients^2)$ per user, and $\mathcal{O}(\dimension)$ for the aggregated share. The cost for the federator is $\mathcal{O}(\nclients+\dimension \nclients)$.
\end{enumerate}

\section{Numerical Evaluation} \label{sec:experiments}

\begin{table}[!t]
\centering
\caption{Communication Complexity. For ByzSecAgg, the partition parameter is in the range $\partition \in [\frac{\nclients+1}{2} - \nbyz - \ncol]$. The partition parameter for ByITFL is a function of the degree $\varphi$ of a polynomial approximation and reads $\partition \in [\frac{\nclients-1}{\varphi+2}-\nbyz-\ncol+1]$.\vspace{-.3cm}}
\begin{tabular}{c|c|c}\label{tab:comm}
    & Per-User & Federator\\ \hline
BREA \cite{so2020byzantine} & $\mathcal{O}(\dimension \nclients+\nclients^2)$ & $\mathcal{O}(\dimension \nclients+\nclients^3)$ \\ \hline
ByzSecAgg \cite{jahani2023byzsecagg} & $\mathcal{O}(\frac{\dimension}{\partition}\nclients+\nclients^2)$ & $\mathcal{O}(\frac{\dimension}{\partition}\nclients+\nclients^3)$ \\ \hline
ByITFL \cite{xia2024byzantine} & $\mathcal{O}(\frac{\nclients}{\partition}\nclients^3+\nclients^4)$ & $\mathcal{O}(\frac{\dimension}{\partition}\nclients+\nclients^2)$  \\\hline
Ours & $\mathcal{O}(\dimension \nclients^2 + \nclients^2)$ & $\mathcal{O}(\dimension \nclients^2 + \nclients^3)$
\end{tabular}
\vspace{-.5cm}
\end{table}

We first compare the communication cost of our scheme to the schemes of \cite{jahani2023byzsecagg,xia2024byzantine,so2020byzantine} in \cref{tab:comm} and observe that our scheme presents a more robust aggregation at the expense of a larger communication cost. 

Then, we evaluate our private and Byzantine resilient framework with NNM in different settings, where data heterogeneity is modeled by a Dirichlet distribution over the clients for each label in the respective dataset. We first consider a simple logistic regression problem on MNIST and a variety of common attacks, i.e., \textit{A little is enough} (ALIE) \cite{baruch2019little}, \textit{Fall of Empires} (FOE) \cite{xie2020fall}, \textit{Sign Flipping} (SF) \cite{allen2021byzantine} and \textit{Label Flipping} (LF) \cite{allen2021byzantine}. For ALIE and FOE, we optimize the attack vector with respect to the underlying robust aggregation rule $\robustaggop$, similar to \cite{allouah2023fixing}. The models are federatedly trained on $\nclients=40$ clients, whereby $\nbyz=10$ are Byzantine, for $400$ epochs with a learning rate $\lr=0.01$ and a Dirichlet parameter of $\beta=0.1$ (strong heterogeneity). We use stochastic quantization with $1024$ quantization levels. We evaluate our method on standard stochastic gradient descent (SGD) and with zero-order (ZO) gradient estimation defined next.
\begin{definition}[Zero-Order Gradient Estimate]
Let $\nseeds\in \mathbb{Z}^+$, a symmetric zero-order estimate $\githat \!\! \define \! \sum_{\seed = 1}^\nseeds \! \githatentry[\seed]\!$ is computed as \vspace{-.3cm}
\begin{align*}
    \githatentry[\seed] = \dimension \frac{\loss(\model\!+\!\dev \perturb, \data) \!-\! \loss(\model\!-\!\dev \perturb, \data)}{2\dev}\perturb \in \mathbb{R}, \seed \in [\nseeds], \vspace{-.6cm}
\end{align*}
where $\dev>0$ is a parameter and the vectors $\perturb, \seed \in [\nseeds],$ called random perturbations\footnote{The $\perturb$'s are equal for all clients and can be sampled through shared randomness to reduce communication costs. We refer to \cite{egger2025byzantine} for more details.} of the model, are sampled uniformly at random from the unit sphere $\mathbb{S}^\dimension = \{\mathbf{x} \in \mathbb{R}^\dimension: \Vert \mathbf{x} \Vert_2 = 1\}$.
\end{definition}

For ZO, we use $\nseeds=64$ and $\dev=0.001$, thereby providing a compression ratio of $\frac{\dimension}{\nseeds} = \frac{7840}{64} \approx 123$. For both cases, we show the results for $\robustaggop$ being \krum (KR) and Multi-\krum (MKR), with and without NNM, respectively. We show in \cref{tab:results} the maximum test accuracies averaged over $5$ runs with different seeds, including their standard deviations. BREA \cite{so2020byzantine} and ByzSecAgg \cite{jahani2023byzsecagg} correspond to SGD-MKR without NNM. ByITFL \cite{xia2024byzantine} is omitted from the comparison as it requires a public root dataset, not required by the other schemes. 
It can be found that NNM improves the Byzantine resilience uniformly over all attacks, for both SGD and ZO, and using \krum and Multi-\krum. We provide in \cref{fig:label_flipping} the accuracy for SF over time, underlining the performance improvements brought by NNM, and in \cref{tab:results} extensive results for all attacks and methods. We stress that our goal is not to optimize the models to yield the best possible accuracies under Byzantine behavior but to provide a comprehensive overview of a large class of attacks and methods.

\setlength{\tabcolsep}{3pt}
 
\begin{table}[!t]
\centering
\caption{Mean and standard deviation of maximum accuracies across $5$ runs with standard deviations for ALIE, FOE, SF and LF attacks for different countermeasures and optimization methods with and without NNM. The performance of BREA \cite{so2020byzantine} and ByzSecAgg \cite{jahani2023byzsecagg} is given by SGD-MKR.\vspace{-.2cm}}
\label{tab:results}
\begin{tabular}{l|c|c|c|c}
\hline
MODE & ALIE \cite{baruch2019little} & FOE \cite{xie2020fall} & SF \cite{allen2021byzantine} & LF \cite{allen2021byzantine}\\
\hline \hline
ZO-KR & 75.3 $\!\pm\!$ 4.3 & 36.9 $\!\pm\!$ 6.2 & 48.6 $\!\pm\!$ 5.9 & 68.7 $\!\pm\!$ 3.4 \\
ZO-KR-NNM & 88.0 $\!\pm\!$ 2.2 & 55.1 $\!\pm\!$ 3.2 & 59.3 $\!\pm\!$ 4.0 & 89.2 $\!\pm\!$ 2.2 \\
SGD-KR & 74.0 $\!\pm\!$ 8.7 & 23.5 $\!\pm\!$ 14.1 & 45.5 $\!\pm\!$ 5.5 & 62.1 $\!\pm\!$ 8.2 \\
SGD-KR-NNM & 86.9 $\!\pm\!$ 1.3 & 52.2 $\!\pm\!$ 4.8 & 61.5 $\!\pm\!$ 5.2 & 89.7 $\!\pm\!$ 1.6 \\
ZO-MKR & 86.5 $\!\pm\!$ 2.1 & 48.4 $\!\pm\!$ 6.4 & 53.6 $\!\pm\!$ 2.6 & 88.4 $\!\pm\!$ 1.2 \\
ZO-MKR-NNM & 89.2 $\!\pm\!$ 0.6 & 54.9 $\!\pm\!$ 5.4 & 60.8 $\!\pm\!$ 2.7 & 90.0 $\!\pm\!$ 0.4 \\
SGD-MKR & 86.1 $\!\pm\!$ 1.0 & 45.5 $\!\pm\!$ 5.7 & 52.8 $\!\pm\!$ 3.1 & 87.1 $\!\pm\!$ 2.2 \\
SGD-MKR-NNM & 86.9 $\!\pm\!$ 0.8 & 60.0 $\!\pm\!$ 4.9 & 61.1 $\!\pm\!$ 5.0 & 89.3 $\!\pm\!$ 2.4 \\
\hline
\end{tabular}
\vspace{-.4cm}
\end{table}

Lastly, we show in \cref{fig:fine_tuning} the results for fine-tuning RoBERTa-large \cite{liu2019roberta} on SST-2 \cite{socher2013recursive}, a dataset for Sentiment analysis, under the ALIE attack. We employ ZO optimization with $\nclients=15$ workers, out of which $\nbyz=3$ are Byzantine, with $\lr=10^{-6}$, a Dirichlet parameter of $\beta=0.1$, and $20000$ epochs. We test our approach against the coordinate-wise trimmed mean (TM) aggregation from \cite{neto2024communication}, here further improved with NNM. We use $\nseeds=5$ random perturbations per epoch and $\dev=0.001$. While trimmed mean cannot be made private, we show that our protocol, allowing the use of private \krum and Multi-\krum with NNM, achieves state-of-the-art accuracies. We observed in simulations not included here for space constraints that, for fine-tuning tasks, the performance gains from Multi-\krum to Multi-\krum with NNM are less significant than for standard training processes such as in \cref{fig:label_flipping}. However, the communication cost compared to SGD is reduced by a factor of $355\cdot 10^6 / 5 = 71\cdot 10^6$, thereby pushing private aggregation methods into practical regimes.

\balance
\bibliographystyle{IEEEtran}
\bibliography{refs}

\begin{thebibliography}{10}
\providecommand{\url}[1]{#1}
\csname url@samestyle\endcsname
\providecommand{\newblock}{\relax}
\providecommand{\bibinfo}[2]{#2}
\providecommand{\BIBentrySTDinterwordspacing}{\spaceskip=0pt\relax}
\providecommand{\BIBentryALTinterwordstretchfactor}{4}
\providecommand{\BIBentryALTinterwordspacing}{\spaceskip=\fontdimen2\font plus
\BIBentryALTinterwordstretchfactor\fontdimen3\font minus \fontdimen4\font\relax}
\providecommand{\BIBforeignlanguage}[2]{{%
\expandafter\ifx\csname l@#1\endcsname\relax
\typeout{** WARNING: IEEEtran.bst: No hyphenation pattern has been}%
\typeout{** loaded for the language `#1'. Using the pattern for}%
\typeout{** the default language instead.}%
\else
\language=\csname l@#1\endcsname
\fi
#2}}
\providecommand{\BIBdecl}{\relax}
\BIBdecl

\bibitem{mcmahan2017communication}
B.~McMahan, E.~Moore, D.~Ramage, S.~Hampson, and B.~A. y~Arcas, ``Communication-efficient learning of deep networks from decentralized data,'' in \emph{Artificial intelligence and statistics}, 2017, pp. 1273--1282.

\bibitem{yin2021comprehensive}
X.~Yin, Y.~Zhu, and J.~Hu, ``A comprehensive survey of privacy-preserving federated learning: A taxonomy, review, and future directions,'' \emph{ACM Computing Surveys (CSUR)}, vol.~54, no.~6, pp. 1--36, 2021.

\bibitem{rodriguez2023survey}
N.~Rodr{\'\i}guez-Barroso, D.~Jim{\'e}nez-L{\'o}pez, M.~V. Luz{\'o}n, F.~Herrera, and E.~Mart{\'\i}nez-C{\'a}mara, ``Survey on federated learning threats: Concepts, taxonomy on attacks and defences, experimental study and challenges,'' \emph{Information Fusion}, vol.~90, pp. 148--173, 2023.

\bibitem{bonawitz2017practical}
K.~Bonawitz, V.~Ivanov, B.~Kreuter, A.~Marcedone, H.~B. McMahan, S.~Patel, D.~Ramage, A.~Segal, and K.~Seth, ``Practical secure aggregation for privacy-preserving machine learning,'' in \emph{ACM SIGSAC Conference on Computer and Communications Security}, 2017, pp. 1175--1191.

\bibitem{bell2020secure}
J.~H. Bell, K.~A. Bonawitz, A.~Gasc\'{o}n, T.~Lepoint, and M.~Raykova, ``Secure single-server aggregation with (poly)logarithmic overhead,'' in \emph{ACM SIGSAC Conference on Computer and Communications Security}, 2020, p. 1253–1269.

\bibitem{so2021turbo}
J.~So, B.~Güler, and A.~S. Avestimehr, ``Turbo-aggregate: Breaking the quadratic aggregation barrier in secure federated learning,'' \emph{IEEE Journal on Selected Areas in Information Theory}, vol.~2, no.~1, pp. 479--489, 2021.

\bibitem{kadhe2020fastsecagg}
S.~Kadhe, N.~Rajaraman, O.~O. Koyluoglu, and K.~Ramchandran, ``Fast{S}ec{A}gg: Scalable secure aggregation for privacy-preserving federated learning,'' \emph{arXiv preprint arXiv:2009.11248}, 2020.

\bibitem{so2022lightsecagg}
J.~So, C.~J. Nolet, C.-S. Yang, S.~Li, Q.~Yu, R.~E.~Ali, B.~Guler, and S.~Avestimehr, ``{LightSecAgg}: a lightweight and versatile design for secure aggregation in federated learning,'' in \emph{Machine Learning and Systems}, vol.~4, 2022, pp. 694--720.

\bibitem{jahani2022swiftagg+}
T.~Jahani-Nezhad, M.~A. Maddah-Ali, S.~Li, and G.~Caire, ``Swiftagg+: Achieving asymptotically optimal communication loads in secure aggregation for federated learning,'' \emph{IEEE Journal on Selected Areas in Communications}, vol.~41, no.~4, pp. 977--989, 2023.

\bibitem{schlegel2023coded}
R.~Schlegel, S.~Kumar, E.~Rosnes, and A.~G.~i. Amat, ``Codedpaddedfl and codedsecagg: Straggler mitigation and secure aggregation in federated learning,'' \emph{IEEE Transactions on Communications}, vol.~71, no.~4, pp. 2013--2027, 2023.

\bibitem{sami2023secure}
H.~U. Sami and B.~Güler, ``Secure aggregation for clustered federated learning,'' in \emph{IEEE International Symposium on Information Theory (ISIT)}, 2023, pp. 186--191.

\bibitem{egger2024private}
M.~Egger, C.~Hofmeister, A.~Wachter-Zeh, and R.~Bitar, ``Private aggregation in hierarchical wireless federated learning with partial and full collusion,'' \emph{arXiv preprint arXiv:2306.14088}, 2024.

\bibitem{blanchard2017machine}
P.~Blanchard, E.~M. El~Mhamdi, R.~Guerraoui, and J.~Stainer, ``Machine learning with adversaries: Byzantine tolerant gradient descent,'' \emph{Advances in neural information processing systems}, vol.~30, 2017.

\bibitem{allen2021byzantine}
Z.~Allen-Zhu, F.~Ebrahimianghazani, J.~Li, and D.~Alistarh, ``Byzantine-resilient non-convex stochastic gradient descent,'' in \emph{International Conference on Learning Representations}, 2021.

\bibitem{chen2017distributed}
Y.~Chen, L.~Su, and J.~Xu, ``Distributed statistical machine learning in adversarial settings: Byzantine gradient descent,'' \emph{ACM on Measurement and Analysis of Computing Systems}, vol.~1, no.~2, pp. 1--25, 2017.

\bibitem{el2021collaborative}
E.~M. El-Mhamdi, S.~Farhadkhani, R.~Guerraoui, A.~Guirguis, L.-N. Hoang, and S.~Rouault, ``Collaborative learning in the jungle (decentralized, byzantine, heterogeneous, asynchronous and nonconvex learning),'' \emph{Advances in neural information processing systems}, vol.~34, pp. 25\,044--25\,057, 2021.

\bibitem{karimireddy2021learning}
S.~P. Karimireddy, L.~He, and M.~Jaggi, ``Learning from history for byzantine robust optimization,'' in \emph{International Conference on Machine Learning}, 2021, pp. 5311--5319.

\bibitem{karimireddy2022byzantinerobust}
------, ``Byzantine-robust learning on heterogeneous datasets via bucketing,'' in \emph{International Conference on Learning Representations}, 2022.

\bibitem{ozfatura2023byzantines}
K.~{\"O}zfatura, E.~{\"O}zfatura, A.~K{\"u}p{\c{c}}{\"u}, and D.~Gunduz, ``Byzantines can also learn from history: Fall of centered clipping in federated learning,'' \emph{IEEE Transactions on Information Forensics and Security}, vol.~19, pp. 2010--2022, 2023.

\bibitem{so2020byzantine}
J.~So, B.~G{\"u}ler, and A.~S. Avestimehr, ``Byzantine-resilient secure federated learning,'' \emph{IEEE Journal on Selected Areas in Communications}, vol.~39, no.~7, pp. 2168--2181, 2020.

\bibitem{shamir1979how}
A.~Shamir, ``How to share a secret,'' \emph{Communications of the ACM}, vol.~22, no.~11, pp. 612--613, 1979.

\bibitem{mceliece1981sharing}
R.~J. McEliece and D.~V. Sarwate, ``On sharing secrets and reed-solomon codes,'' \emph{Communications of the ACM}, vol.~24, no.~9, pp. 583--584, 1981.

\bibitem{jahani2023byzsecagg}
T.~Jahani-Nezhad, M.~A. Maddah-Ali, and G.~Caire, ``Byzsecagg: A byzantine-resistant secure aggregation scheme for federated learning based on coded computing and vector commitment,'' \emph{arXiv preprint arXiv:2302.09913}, 2023.

\bibitem{hou2024priroagg}
S.~Hou, S.~Li, T.~Jahani-Nezhad, and G.~Caire, ``Priroagg: Achieving robust model aggregation with minimum privacy leakage for federated learning,'' \emph{arXiv preprint arXiv:2407.08954}, 2024.

\bibitem{pillutla2022robust}
K.~Pillutla, S.~M. Kakade, and Z.~Harchaoui, ``Robust aggregation for federated learning,'' \emph{IEEE Transactions on Signal Processing}, vol.~70, pp. 1142--1154, 2022.

\bibitem{ozdayi2021defending}
M.~S. Ozdayi, M.~Kantarcioglu, and Y.~R. Gel, ``Defending against backdoors in federated learning with robust learning rate,'' in \emph{AAAI Conference on Artificial Intelligence}, vol.~35, no.~10, 2021, pp. 9268--9276.

\bibitem{xia2024byzantine}
Y.~Xia, C.~Hofmeister, M.~Egger, and R.~Bitar, ``Byzantine-resilient secure aggregation for federated learning without privacy compromises,'' in \emph{IEEE Information Theory Workshop (ITW)}, 2024, pp. 223--228.

\bibitem{cao2020fltrust}
X.~Cao, M.~Fang, J.~Liu, and N.~Z. Gong, ``Fltrust: Byzantine-robust federated learning via trust bootstrapping,'' \emph{arXiv preprint arXiv:2012.13995}, 2020.

\bibitem{ben2019completeness}
M.~Ben-Or, S.~Goldwasser, and A.~Wigderson, ``Completeness theorems for non-cryptographic fault-tolerant distributed computation,'' in \emph{ACM Symposium on Theory of Computing}, 1988, p. 1–10.

\bibitem{xia2024lobyitfl}
Y.~Xia, C.~Hofmeister, M.~Egger, and R.~Bitar, ``Lobyitfl: Low communication secure and private federated learning,'' \emph{arXiv preprint arXiv:2405.19217}, 2024.

\bibitem{charikar2017learning}
M.~Charikar, J.~Steinhardt, and G.~Valiant, ``Learning from untrusted data,'' in \emph{Annual ACM SIGACT Symposium on Theory of Computing}, 2017, pp. 47--60.

\bibitem{liu2021approximate}
S.~Liu, N.~Gupta, and N.~H. Vaidya, ``Approximate byzantine fault-tolerance in distributed optimization,'' in \emph{ACM Symposium on Principles of Distributed Computing}, 2021, pp. 379--389.

\bibitem{allouah2023fixing}
Y.~Allouah, S.~Farhadkhani, R.~Guerraoui, N.~Gupta, R.~Pinot, and J.~Stephan, ``Fixing by mixing: A recipe for optimal byzantine ml under heterogeneity,'' in \emph{International Conference on Artificial Intelligence and Statistics}, 2023, pp. 1232--1300.

\bibitem{malladi2023fine}
S.~Malladi, T.~Gao, E.~Nichani, A.~Damian, J.~D. Lee, D.~Chen, and S.~Arora, ``Fine-tuning language models with just forward passes,'' \emph{Advances in Neural Information Processing Systems}, vol.~36, pp. 53\,038--53\,075, 2023.

\bibitem{neto2024communication}
A.~d. S.~D. Neto, M.~Egger, M.~Bakshi, and R.~Bitar, ``Communication-efficient byzantine-resilient federated zero-order optimization,'' \emph{arXiv preprint arXiv:2406.14362}, 2024.

\bibitem{egger2025byzantine}
M.~Egger, M.~Bakshi, and R.~Bitar, ``Byzantine-resilient zero-order optimization for communication-efficient heterogeneous federated learning,'' \emph{arXiv preprint arXiv:2502.00193}, 2025.

\bibitem{gupta2015deep}
S.~Gupta, A.~Agrawal, K.~Gopalakrishnan, and P.~Narayanan, ``Deep learning with limited numerical precision,'' in \emph{International conference on machine learning}, 2015, pp. 1737--1746.

\bibitem{croci2022stochastic}
M.~Croci, M.~Fasi, N.~J. Higham, T.~Mary, and M.~Mikaitis, ``Stochastic rounding: implementation, error analysis and applications,'' \emph{Royal Society Open Science}, vol.~9, no.~3, p. 211631, 2022.

\bibitem{yin2018byzantine}
D.~Yin, Y.~Chen, R.~Kannan, and P.~Bartlett, ``Byzantine-robust distributed learning: Towards optimal statistical rates,'' in \emph{International conference on machine learning}, 2018, pp. 5650--5659.

\bibitem{ngo2024secure}
K.-H. Ngo, J.~{\"O}stman, G.~Durisi, and A.~Graell~i Amat, ``Secure aggregation is not private against membership inference attacks,'' in \emph{Joint European Conference on Machine Learning and Knowledge Discovery in Databases}, 2024, pp. 180--198.

\bibitem{tajeddine2019private}
R.~Tajeddine, O.~W. Gnilke, D.~Karpuk, R.~Freij-Hollanti, and C.~Hollanti, ``Private information retrieval from coded storage systems with colluding, byzantine, and unresponsive servers,'' \emph{IEEE Transactions on information theory}, vol.~65, no.~6, pp. 3898--3906, 2019.

\bibitem{baruch2019little}
G.~Baruch, M.~Baruch, and Y.~Goldberg, ``A little is enough: Circumventing defenses for distributed learning,'' \emph{Advances in Neural Information Processing Systems}, vol.~32, 2019.

\bibitem{xie2020fall}
C.~Xie, O.~Koyejo, and I.~Gupta, ``Fall of empires: Breaking byzantine-tolerant sgd by inner product manipulation,'' in \emph{Uncertainty in Artificial Intelligence}, 2020, pp. 261--270.

\bibitem{liu2019roberta}
Y.~Liu, ``Roberta: A robustly optimized bert pretraining approach,'' \emph{arXiv preprint arXiv:1907.11692}, vol. 364, 2019.

\bibitem{socher2013recursive}
R.~Socher, A.~Perelygin, J.~Wu, J.~Chuang, C.~D. Manning, A.~Y. Ng, and C.~Potts, ``Recursive deep models for semantic compositionality over a sentiment treebank,'' in \emph{Conference on empirical methods in natural language processing}, 2013, pp. 1631--1642.

\end{thebibliography}

\end{document}